\pdfoutput=1

\documentclass[11pt]{article}

\usepackage[final]{acl}

\usepackage{times}
\usepackage{latexsym}

\usepackage[T1]{fontenc}

\usepackage[utf8]{inputenc}

\usepackage{microtype}

\usepackage{inconsolata}
\usepackage{enumitem}

\usepackage{graphicx}

\usepackage{amsmath}
\usepackage{amssymb}
\usepackage{booktabs}
\usepackage{soul}

\title{Generating Faithful and Salient Text from Multimodal Data}

\author{Tahsina Hashem$^1$, Weiqing Wang$^1$, Derry Tanti Wijaya$^2$, \\ \textbf{Mohammed Eunus Ali}$^3$, \textbf{Yuan-Fang Li}$^1$  \\ 
  $^1$Department of Data Science \& AI, Monash University, Australia \\ $^2$Department of Data Science, Monash University, Indonesia \\ $^3$Department of CSE, Bangladesh University of Engineering and Technology, Bangladesh  \\
  \small{\texttt{\{tahsina.hashem, Teresa.Wang, derry.wijaya, yuanfang.li\}@monash.edu}}; \\ \small{\texttt{eunus@cse.buet.ac.bd}} \\
  }

\begin{document}
\maketitle
\begin{abstract}


While large multimodal models (LMMs) have obtained strong performance on many multimodal tasks, they may still hallucinate while generating text. Their performance on detecting salient features from visual data is also unclear. 
In this paper, we develop a framework to generate faithful and salient text from mixed-modal data, which includes images and structured data ( represented in knowledge graphs or tables). 
Specifically, we train a small \emph{vision critic model} to identify hallucinated and non-salient features from the image modality. The critic model also generates a list of salient image features. This information is used in the \emph{post editing} step to improve the generation quality. 
Experiments on two datasets show that our framework improves LMMs' generation quality on both faithfulness and saliency, outperforming recent techniques aimed at reducing hallucination. The dataset and code are available at \href{https://github.com/TahsinaHashem/FaithD2T}{https://github.com/TahsinaHashem/FaithD2T}.
\end{abstract}

\section{Introduction}
In many real-world scenarios, data is presented in mixed modalities, in which complementary information is contained. Examples include product brochures, scientific/technical publications, and news articles.  
Structured data-to-text generation is the task of generating natural language sentences from the data in a structured format, such as tables, knowledge graphs, or databases. Researchers have proposed several models to 
make this structured information more accessible
to humans, aiming to generate fluent, informative,
and faithful text descriptions or summaries from the structured data. This task has a wide range of applications across different industries and domains i.e. house advertising, financial reporting, automated journalism, medical reporting,  e-commerce product descriptions, generating biographies, etc.

Significant progress has been made in data-to-text generation tasks. 
Several well-known models have utilized pre-trained language models (PLMs) such as
BART~\cite{lewis2019bart}, T5~\cite{T5raffel2020exploring} or GPT~\cite{GPTradford2019language} with appropriate structure-aware frameworks  ~\cite{colas2022gap, han2022selfGraphtoText,li2024unifyingGraphtoTextModelLinear} to generate text descriptions from the structured data. However, the importance of multimodal input with structured data was not extensively addressed. The problem was explored on a small scale by~\cite{gatti2022vistot,yang2023multimodaldatatotext}. Their proposed model aimed to generate a one-line summary sentence from a given table and an associated image. They showed that integrating vision data with structured data would lead to more informative and relevant text. However, the research did not consider their generation task's saliency and faithfulness.

Recently, several open-sourced large multimodal models (LMMs)~\cite{liu2023improvedLLAVA1.5,zhu2023minigpt,instructblip} show promising performance in a variety of multimodal tasks~\cite{bai2023qwen, liu2024visual,lu2022crossmodalRetrieval, yin2023survey,gupta2023cliptrans} i.e. image captioning, visual question answering, multimodal conversation, cross-modal retrieval, etc. In this research, we exploit these powerful LMMs to generate text from structured data (knowledge graph and table) with images. We have examined the performance of two prominent LMMs, LLaVA-1.5~\cite{liu2023improvedLLAVA1.5} and MiniGPT4~\cite{zhu2023minigpt} on two advertising multimodal (structured data with images) datasets i.e. the real-estate house dataset~\cite{das2021boostingHouse} and the e-commerce product dataset~\citep{shao2019longProductData}. The models generate good-quality advertising text but have two types of limitations: (1) generate some hallucinated information
that is not aligned with the vision input; (2) unable to detect salient image features. These limitations hamper the faithfulness and saliency of the generated text.

\begin{figure*}[htb]
\begin{center}
\includegraphics[width=0.9\textwidth]{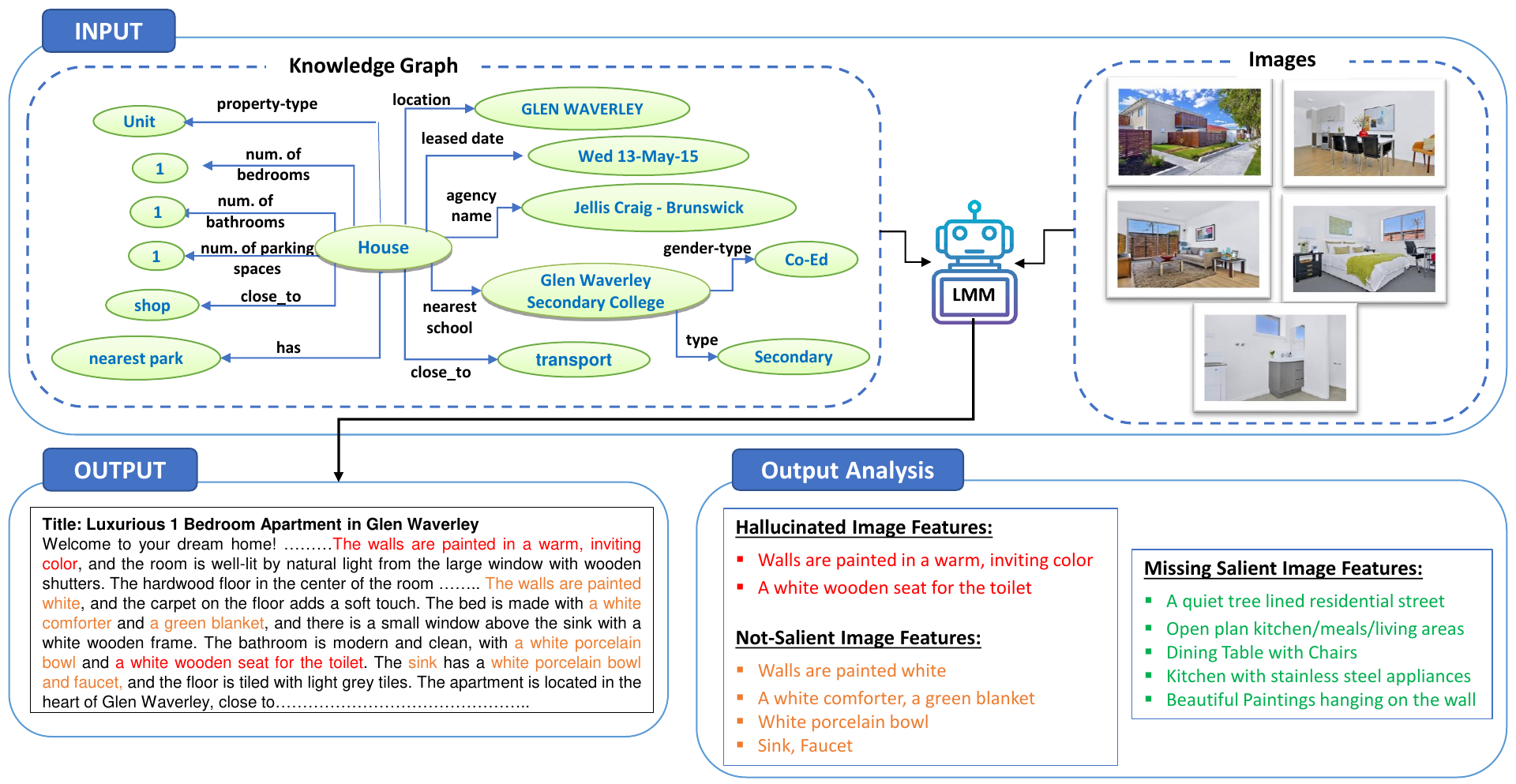}
\end{center}
\caption{A Sample Input and Output of an LMM: MiniGPT4. The Output Analysis lists the errors.}
\label{fig:introFig}
\vspace{-1.5em}
\end{figure*}
Figure~\ref{fig:introFig} shows an example input and output of an LMM: MiniGPT4. The input consists of a small KG about a house, which contains information on its internal features and neighborhood, and the corresponding images of the house, which gives a detailed outlook of the properties, from a real-world real-estate KG~\citep{das2021boostingHouse}. The output shows the text generated by the LMM. The generated text describes the graph features of the house accurately but struggles to describe the image features accurately. The Output Analysis lists some limitations of LMM: The LMM mentions some features (i.e.\ hallucination, highlighted in \textcolor{red}{red}) that are not aligned with the input images. The LMM also lists some features (i.e.\ not-salient features, highlighted in \textcolor{orange}{orange}) that deteriorate the saliency of the generated text while missing some features in the ground-truth text (i.e.\ salient features, highlighted in \textcolor{green}{green}) that are important to make the text attractive for advertising purposes.

Visual hallucination problems of LMMs cause a serious negative impact on visual-to-text generation and reasoning tasks ~\citep{liu2023aligningTuning,wang2023evaluationLVLM, gunjal2024detectingHalNoReward, jing2023faithscore}. Researchers have started proposing different strategies~\cite{liu2023aligningTuning,wang2024vigc,sun2023aligningReward,zhou2024analyzingLURE,yin2023woodpeckerHalCorrection} to reduce object hallucinations. Most of the techniques are based on instruction-tuning~\citep{liu2023aligningTuning,liu2023mitigatingHalTuning} or filtering the hallucination information from the training data~\citep{wang2024vigc,yu2023hallucidoctor} and then fine-tuning the models with the revised version of the dataset. This process of preparing such a good number of high-quality instructions or datasets is time-consuming and costly. Some researchers~\cite {sun2023aligningReward} have utilized reinforcement learning from human
feedback in training the LMMs using reward models. Another alternate way of mitigating hallucination is post-hoc detection and correction frameworks~\citep{zhou2024analyzingLURE,yin2023woodpeckerHalCorrection}. These methods are cost-friendly and showed good performance in mitigating hallucinations in the generated test. 

Our proposed framework follows the detection and correction strategy but instead of fine-tuning LMMs, we train a small vision language model (VLM)~\citep{li2023blip2} as a transparent vision critic model that can detect the errors of the text generated by LMMs with an explanation and list the missing salient image features of the text generated by LMMs. Finally, we update the generated text using LLM from the feedback of the critic model using an appropriate prompt.

The contributions of our research work are:
\begin{itemize}[noitemsep,nolistsep]
    \item Propose a novel task of generating faithful and salient natural language text from structured data and images.
    \item Design a framework to train a small vision model to act like an interpretable vision critic model that can verify the faithfulness and saliency of the features as well as list the missing salient image features of the text generated by LMMs. 
     \item Experimental Results demonstrate the effectiveness of our model over existing baselines.
\end{itemize}



\section{Related Work}

\subsection{Multimodal Data to Text generation}
Several structure-enhanced pre-trained language models~\cite{han2022selfGraphtoText,li2024unifyingGraphtoTextModelLinear,tang2023mvpTabletoText,liu2022plogTabletoText} showed good performance in structured data-to-text generation tasks. However, very few works~\cite{gatti2022vistot,yang2023multimodaldatatotext} have been done in multimodal data-to-text generation tasks. An initial attempt was made by Gatti et al.~\cite{gatti2022vistot} to generate a one-line summary sentence from vision-augmented tabular data. They proposed a VT3 multimodal transformer that consists of a BART model~\cite{lewis2019bart} and a vision transformer~\citep{liu2021swintransformer}, that can generate text auto-regressively. A different approach  was proposed 
to overcome a large amount of annotated training data requirement ~\cite{yang2023multimodaldatatotext}. They proposed a multimodal prompt learning framework to accurately generate titles for novel products with limited labels. However, both models aim to generate a one-line summary sentence. They cover a small number of vision features without verifying the saliency and faithfulness of their generated text. Whereas, in our problem, we focus on generating a long advertising text that should contain all the salient and faithful features of the vision data. 

Recently, the large multimodal models~\citep{zhu2023minigpt,liu2023improvedLLAVA1.5,instructblip,ye2023mplug} have shown remarkable success in various multimodal tasks such as image captioning~\citep{lin2014microsoftImageCaptioning}, visual question-answering (VQA)~\citep{antol2015vqa} and multimodal conversation~\citep{liu2024llavaNext}. Hence, our research work exploits these powerful LMMs to generate salient and faithful text from multimodal data.
\subsection{Hallucination in LMMs}
Although LMMs demonstrate strong performance across multiple benchmark tasks and produce quality results, they struggle with the problem of visual hallucination. This issue occurs when the generated responses do not align with the visual input. Researchers investigated this phenomenon in the realm of object hallucination~\cite{li2023evaluatinghalstudy,liu2023mitigatinghalstudy,biten2022lethalstudy}, where the generated content features objects that do not match or are not present in the input image. Recently, it has been shown~\citep{zhai2023halstudy} that this multimodal hallucination happens because the vision encoder does not accurately ground images. They tend to depend more on their built-in knowledge rather than the visual input provided. Furthermore, empirical studies by Wang et al. \citep{wang2023halstudy}, have shown that these models focus more on previously generated tokens than on the image features.

\subsection{Hallucination Mitigation of LMMs}
Researchers have already proposed a number of alternative strategies to minimize the visual hallucination problem of LMMs. Some focus on improving the quality of instruction tuning data. LRV-Instruction dataset~\citep{liu2023mitigatinghalstudy},  VIGC~\citep{wang2024vigc}, M-HalDetect~\citep{gunjal2024detectingFDPO} are examples of such high-quality prepared datasets. Some tried to refine the model training techniques like reinforcement learning from human feedback (RLHF) in LLaVA-RLHF~\citep{sun2023aligningLLaVA-RLHF}, or optimization models in FDPO~\citep{gunjal2024detectingFDPO}. Some researchers apply post hoc detection and correction strategies such as LURE~\citep{zhou2024analyzingLURE} that is based on object co-occurrence, uncertainty, and position in text; and Woodpecker~\citep{yin2023woodpeckerHalCorrection} that extracts key concepts and validates the visual knowledge using object detector and VQA model.

For a more cost-effective approach, 
we adopt the post-hoc detection and correction approach. We train a small pre-trained VLM that can be used in cooperation with LLM to mitigate both the visual hallucinated features and non-salient features.

\section{Method}
To generate high-quality text, finetuning LMM may not be feasible for proprietary models, and it may not be practical due to the prohibitively high resource and data requirements.
Thus, we propose a cost-effective post-hoc detection and correction approach that trains a small VLM to act as a \emph{vision critic model} that identifies errors in the LMM-generated text. With the feedback provided by the critic model, a capable LLM (such as GPT-3.5) is then employed to update the text using this feedback. 

\begin{figure*}[htb]
\begin{center}
\includegraphics[width=0.9\textwidth]{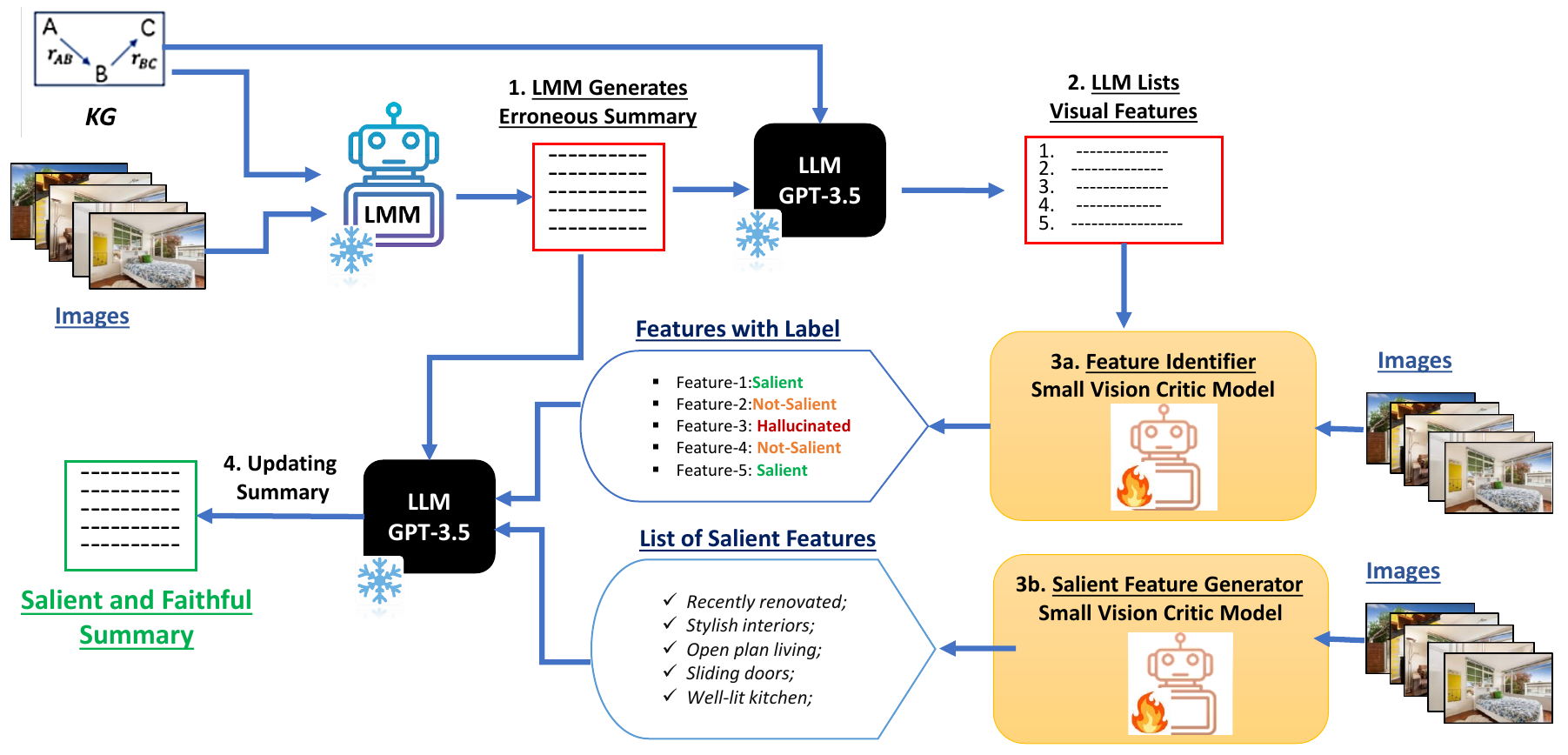}
\end{center}
\caption{The Pipeline of our Framework for Salient and Faithful Multimodal Data to Text Generation 1) Generating Text using LMM 2) Extracting Image Features from the Text using GPT-3.5 3) Trained Vision Critic Model gives feedback to LMM 4) LMM update the Text by making corrections.}
\label{fig:blockdiagram}
\vspace{-1.5em}
\end{figure*}

Figure~\ref{fig:blockdiagram} depicts the overall architecture of our proposed method. Given the text generated from the mixed-modal data by the LMM, we first prompt a capable LLM (such as GPT-3.5) to extract the list of image features from the text by filtering out features from the structured data. With this list and the images as input, we employ our trained vision critic model to identify hallucinated image features and non-salient image features in the text. The critic model also generates salient image features that are missing from the text. Finally, we prompt the LLM to remove the hallucinated and non-salient features from the text and append the missing salient image features to the text.

\subsection{Problem Formulation}

Given a training dataset $\mathcal{D}=(X, Y)$, in which $X=\{(s_1,i_1),(s_2,i_i),\ldots,(s_{|\mathcal{D}|},i_{|\mathcal{D}|})\}$ is a mixed-modal dataset that consists of pairs of structured data $s_i$ (i.e.\ knowledge graphs or tables) and (multiple) images $i_i$, and $Y=\{y_1,\ldots,y_{|\mathcal{D}|}\}$ is a set of reference text for each $x_i$, our aim is to train a model that generates a text passage $\hat{y}_j$ for $x_j = (s_j, i_j)$ that is both faithful to $s_j$ and $i_j$ and contains the salient image features in $i_j$. Note that $Y$ may contain hallucinated information. 

We assume the structured data is either a knowledge graph or a table. Let $KG$ = $(V, E)$ represent a knowledge graph, where $V=\{{e_1}, {e_2}, \dots, {e_{|V|}} \}$ represents the entity set and $E = \{r_{ij}\} \subseteq V\times V$ represents the relations connecting the entities. For the tabular data, let $T=\{({a_1},{v_1}), ({a_2},{v_2}), \dots, ({a_m},{v_m}) \}$  represents a table with $m$ number of attribute-value pairs. Every type of structured data contains an image set. Let $I=\{{i_1}, {i_2}, \dots, {i_l} \}$ represents the corresponding image set. 



\subsection{Training  a Small Vision Language Model}
We choose a small vision language model (VLM) BLIP-2~\cite{li2023blip2} to act as a critic model. 
BLIP-2 addresses the modality gap by employing a lightweight Querying Transformer (Q-Former). 
BLIP-2 utilizes a generic and efficient pretraining strategy that bootstraps vision-language pretraining from off-the-shelf frozen pretrained image encoders and frozen large language models (LLMs). It shows good performance on visual question-answering tasks, image captioning tasks, image-text retrieval tasks and visual commonsense reasoning tasks~\cite{park2024localizedBlip2Critic}. 

Recently, it has been shown~\citep{kimparameterPEFTApplication} that Parameter-efficient fine-tuning (PEFT)~\citep{peft} maintains competitive performance while requiring much less computational memory. Thus, we apply LoRA~\citep{hu2021lora} to the Q-Former and the base LLMs, Flan-T5-XL, of the BLIP-2 model. This allows us to fine-tune the BLIP-2 model in a cost-effective way.

In the following two subsections we discuss our training process in detail for the two critic tasks:

\subsubsection{Classifying Image Feature}
We observe that LMMs cannot reliably distinguish salient features from non-salient features and hallucinated features, degrading the quality of generated text. 
Thus, we train the vision critic model to become an expert in detecting whether a feature is salient, non-salient or hallucinated given an input image. 
We formulate this task as a generation problem, where a set $x = (i, f)$ is given to BLIP-2 vision critic model, with $i$ being an image and $f$ being a feature, and the output is a textual output of the label $y\in\{salient, non\text{-}salient, hallucinated\}$ with an appropriate explanation. We use a standard conditional language modeling loss function:
\begin{equation}\label{eq3.2}
        L_{CE} = - \sum_{i = 1}^n log P (y_i|y_{<i}, X)
\end{equation}
Our training data consists of a set of labeled image-feature pairs along with the corresponding rationales for the three categories. 

\subsubsection{Listing Salient Image Features}
We train our vision critic model to identify the important i.e.\ salient features of a given image. We formulate this task also as a generation problem, where the vision critic model outputs a list of salient features, $S_i=\{[{s_1}]; [{s_2}], \ldots; [{s_m}] \}$ given an image $i$.  
We fine-tune the vision critic model by maximizing the log-likelihood:
\\
\begin{equation}\label{eq3.3}
        L_{S_i} = - \mathbb{E}_{(I,S_i) \sim \mathcal{D'}} \log P (S_i| i)
\end{equation}
Here, the training dataset $D' = {(i, S_i)}$ consists of an image and a list of salient image features. The training data generation process is discussed in detail below.

\subsubsection{Training Data Generation}\label{datageneration} 
We prepare labeled data (i.e., image features labeled with salient, non-salient, and  hallucinated) for training the critic model to classify the image features. To generate this data, we take samples of ground-truth texts and the corresponding LMMs-generated texts. We also prepare image-features pairs where each pair is a list of salient features for the corresponding image. This data is used to train the critic model to generate salient features of an image. The entire training data generation process involves the following steps:\\ 
\textbf{(1) Extracting features from text:} Both the ground-truth text and the generated text contain features from the structured data and the images in an aggregated form. 
We use an LLM, i.e.\ GPT-3.5, to list the features one by one from every sentence of the texts following some in-context examples. An example prompt can be found in Figure:~\ref{fig:PromptListOfFeatures}\\
\textbf{(2) Listing visible and non-visible features:}
Both the ground-truth text and the LMM-generated text contain hallucinated information. To prepare labeled visible (i.e., salient or non-salient) and not visible (i.e., potentially hallucinated) features from the images for training the critic model, we prompt GPT-4V with input images and the list of extracted features from (1) to verify whether the feature is visible or not visible in the input images.\\  
\textbf{(3) Listing hallucinated features:}
We input GPT-3.5 the structured data and the list of not visible features that we obtain from (2) and prompt it to list the features not aligned with the structured data. Features that are not aligned with the structured data will be labeled as hallucinated features since they are neither visible in the image nor exists in the structured data.\\
\textbf{(4) Listing salient and non-salient features:} 
We ask GPT-3.5 to compare the visible image features in the LMM-generated text with the ground-truth visible image features. All visible features from the ground-truth text are salient image features. Visible features in the generated text that are similar to any of the features mentioned in the ground-truth text or the structured data are also salient features. The remaining visible features in the LMM-generated text are the non-salient features. \\
\textbf{(5) Generating rationale for feature labels: } \\
After preparing the labeled image-features pairs for salient, non-salient, and hallucinated categories, we prompt GPT-3.5 to generate a one-sentence explanation for why a feature might be labeled salient or not salient.
For the hallucinated feature, we use the default explanation that: "The feature is not visible in the image". These rationales make our vision critic model interpretable and leads to improve the accuracy of the critic model in feature labeling tasks. \\
All the prompt templates are shown in appendix~\ref{app-prompt-design-train}.
\subsection{Post-hoc Text Editing from the Feedback given by the Critic Model}
We design an appropriate prompt to utilize an LLM (GPT-$3.5$) for updating the LMM-generated text according to the feedback of the trained vision critic model. The update operation is done in two steps. Firstly, non-salient and hallucinated image features are pruned from the text. Secondly, salient image features are appended to the pruned text. Figure~\ref{fig:PromptUpdatesummary} shows the prompt template in the Appendix.

\section{Experiments}

\subsection{Dataset}
We conduct experiments and evaluation on two multimodal data-to-text generation datasets: the House dataset of real-estate house listings~\citep{das2021boostingHouse}, containing images and knowledge graphs; and the Product dataset of Chinese e-commerce clothing products~\cite{shao2019longProductData}, containing images and attribute-value information in a tabular format. In both datasets, the ground-truth text contains a significant amount of hallucinated information, making the task of generating faithful text especially challenging. Table~\ref{tab:stat} shows brief statistics of these two datasets. 

\begin{table}[hbtp]
  \centering
  \caption{Statistics of the House and Product Datasets.}
  \label{tab:stat}
  \begin{tabular}{lccc}
    \toprule \\ [-0.98em]
    \textbf{Dataset}& \textbf{Avg.} &\textbf{Avg.} & \textbf{Avg.} \\
    & \textbf{\# triples} &\textbf{\# images} & \textbf{Text length} \\
    \midrule \\[-1.2em]
    House     & 22.8   & 3  & 153       \\
    Product     & 7.4   &  1  & 110      \\\bottomrule
  \end{tabular}
\end{table}

\noindent\textbf{The House dataset}
is a large real-estate and point-of-interests (POI) dataset of Melbourne, Australia~\citep{das2021boostingHouse}. 
It includes 53,220 records of house sales transactions from 2013 to 2015. It consists of three types of POIs, namely regions, schools, and train stations, along with their corresponding features. Each sample in the dataset includes (1) a ground-truth advertisement text, (2) a KG describing house and POI features, and (3) multiple images of the house. However, the given ground-truth text contains a significant level of hallucinated information.
We use $3,100$ samples for training the vision critic model and $100$ test samples for testing the performance of the critic model. We prepared labeled image-feature pairs according to section \ref{datageneration}. 
Details of the training data-split ratio are shown in Appendix~\ref{app:data-split}.

\noindent\textbf{The Product dataset} 
is from a Chinese e-commerce platform of clothings, consisting of 119K samples of advertising text, a clothing specification table, and a single image of the clothing. Each table is a set of attribute-value pairs describing a piece of clothing. The ground-truth advertising text also contains hallucinated information. For training of the critic model, we have used $4,700$ samples and for testing, we used $340$ samples. We prepared labeled image-feature pairs according to section \ref{datageneration}. 
Details of the training data-split ratio are shown in Appendix~\ref{app:data-split}.
\begin{table*}[hbtp]
\small
  \centering
  \caption{Main results on the House dataset. \textbf{Bold} font denotes the best results for each backbone model.}
  \label{tab:HouseResult}
  \begin{tabular}{lccccc}
    \toprule
    \textbf{Model} & \multicolumn{4}{c}{Saliency} & Faithfulness\\
    \cmidrule(lr){2-5} \cmidrule(lr){6-6}
     & BLEU &METEOR & ROUGE-L &BERTScore & CLIP Score\\
    \midrule \\[-1.2em]
    \multicolumn{6}{l}{\textbf{Baseline Model}} \\
    MiniGPT4      & 8.08    & 25.28    & 13.44 &83.98 &  23.92    \\
    MiniGPT4-Woodpecker      & 8.15    & 27.13    & 13.34 &83.91 &  23.89    \\
     MiniGPT4-LURE      & 11.01    & 16.63    & 14.44 & 84.33 &  23.86    \\\midrule \\[-1.2em]
    \multicolumn{6}{l}{\textbf{Our Model}} \\
    MiniGPT4-Pruned   &10.13   & 23.14    & 15.13    & 84.71  &  24.30     \\
    MiniGPT4-Appended   &10.69   & \textbf{28.08}    & 15.58    & 85.15  &  24.26     \\
    MiniGPT4-Combined  &\textbf{11.98}    & 26.09    & \textbf{16.50}    & \textbf{85.36 } &  \textbf{24.59}     \\ \midrule \\[-1.2em]
    \multicolumn{6}{l}{\textbf{Baseline Model}} \\
    LLaVA-1.5      & 11.34    & 29.82   & 16.06 &85.10 &  23.92     \\
    LLaVA-1.5-Woodpecker      & 9.81    & 29.93   & 14.86 &84.66 &  23.89    \\\midrule \\[-1.2em]
    \multicolumn{6}{l}{\textbf{Our Model}} \\
    LLaVA-1.5-Pruned      & 13.74    & 27.36    & 16.87 &85.67 &  24.29     \\
    LLaVA-1.5-Appended      & 13.52    & \textbf{31.65}    & \textbf{17.35} &85.72 &  24.49     \\
    LLaVA-1.5-Combined      & \textbf{15.01}    & 29.29   & 17.33 &\textbf{86.01} &  \textbf{24.63}   \\
    \bottomrule
  \end{tabular}
\end{table*}

\subsection{Baseline Models}
Two prominent LMMs, namely MiniGPT-4~\citep{zhu2023minigpt} and LLaVA-1.5~\citep{liu2023improvedLLAVA1.5} are used as the baseline models. We also compare with two recent post-hoc hallucination detection and correction models, LURE~\citep{zhou2024analyzingLURE} and Woodpecker~\citep{yin2023woodpeckerHalCorrection}. Due to resource constraints, we only experiment with LURE on MiniGPT-4.
The backbone model of LURE is MiniGPT-4. 
Woodpecker utilizes GPT-3.5-turbo as its corrector, grounding
DINO~\citep{liu2023groundingDino} as its object detector and BLIP-2-FlanT5-XXL~\citep{li2023blip2} as its visual question answering model.  

\subsection{Preliminary Analysis}

We conducted a preliminary analysis of the performance of MiniGPT-4 and LLaVA-1.5 with the $100$ test samples of the House and $340$ test samples of the product datasets. For each model, we input the structured data and images and prompt it to generate an advertising text passage. The structured data (KG or table) is given in a linearized format for a better understanding by the LMM~\citep{li2024unifyingGraphtoTextModelLinear}. As LMMs are unable to accept as input multiple images simultaneously, for the House dataset, we input images one by one and ask the LMM to list the key features of the input image. 
Detailed prompt templates are shown inref~\ref{app-prompt-design-gen} in the supplementary files. We observe that the LMMs can accurately list features from the structured data, but struggle to list the image features correctly.

The following common errors are observed in the generated texts by the LMMs:
\begin{itemize}[noitemsep,nolistsep]
    \item \textbf{Missing salient image features}: LMMs sometimes miss some important image features in the generated text that are essential for advertising purposes. We consider the image features listed in the ground-truth text as the standard salient image features. 
    \item \textbf{Hallucinated image features}: LMM-generated text sometimes contain image features that are not present in input images.
    \item \textbf{Non-salient image features}: LMMs sometimes mention features from the images that are not attractive to customers. These features deteriorate the saliency of the text. 
\end{itemize}
Figure~\ref{fig:introFig} shows the text generated by MiniGPT4 from a sample in the House dataset. The output analysis lists the erroneous features (i.e., hallucinated or not salient) in the text as well as the missing salient image features.

\subsection{Experimental Settings}
In our framework, we keep the LMM and the LLM frozen. We finetune the small vision language model Blip-2 ~\cite{li2023blip2}. We apply PEFT fine-tuning ~\citep{peft} to the BLIP-2-FlanT5-XL model~\citep{li2023blip2}. We apply LoRA~\citep{hu2021lora} to both the Q-Former and the base LLMs, Flan-T5-XL. For the House dataset,
we fine-tune both critic models (for feature classification and for generating missing salient features respectively) for $25$ epochs. For the Product dataset, we fine-tune both critic models for $50$ epochs. The batch size for both datasets is set to $16$. The maximum
length of the output text sequences is set to $350$ tokens for the House dataset and $200$ tokens for the Product dataset. We adopt Adam~\citep{kingma2014adam} as the optimizer and set the learning rate to be $5e$-$5$. We used one A40 48GB GPU for all the experiments.

\subsection{Main Results}
Our main experiment aims at the faithfulness and saliency of the text generated by LMMs from the mixed-modal data. 
For saliency evaluation, we consider the image features contained in the ground-truth text as ground-truth salient features. For faithfulness evaluation, we need to pre-process the text to obtain faithful features as the ground-truth text contains hallucinated information. Specifically, we prompt GPT-3.5 to list features from ground-truth text, and prompt GPT-4V to remove hallucinated features from this list (i.e., features that are neither visible in the image nor exist in the structured data). Finally, we prompt GPT-3.5 to generate a paragraph containing the faithful features. Note that, the faithful features in the ground-truth text are also salient. Thus, in this way, we obtain the salient and faithful ground-truth text. The prompt template can be found in the appendix. 

We use automatic metrics to measure both faithfulness and saliency of generated text. 
We employ standard metrics  BLEU~\citep{BLUEpapineni2002bleu}, METEOR~\citep{METEORbanerjee2005meteor}, ROUGE-L~\citep{ROUGElin2004rouge} and BERTScore~\citep{zhang2019bertscore} to measure saliency of the generated text by comparing it with the pre-processed ground-truth text. 
To verify the faithfulness of the generated text with respect to the input image(s), we utilize the CLIP score~\citep{hessel2021clipscore}, which is widely used~\citep{zhou2024analyzingLURE,jing2023faithscore} to measures text-image alignment.

Table~\ref{tab:HouseResult} and Table~\ref{tab:productResult} present the results on the House and Product datasets respectively. 
We evaluate our method, denoted ``\textbf{-Combined}'', against the baselines and other post-hoc hallucination detection and correction models (LURE and Woodpecker). 
From the results of both datasets, we can observe that our method achieves the best performance, outperforming the baseline models and other hallucination-reduction techniques on most settings. Specifically, our method not only outperforms Woodpecker and LURE in reducing hallucination (i.e.\ improving faithfulness), it also achieves the best result in preserving saliency. 

Some qualitative examples of pre-processed ground-truth text and the text generated by different models can be found in Appendix~\ref{sec:samples}.

\begin{table*}[hbtp]
\small
  \centering
  \caption{Main Results on the Product dataset. \textbf{Bold} font denotes the best results for each backbone model.}
  \label{tab:productResult}
  \begin{tabular}{lccccc}
    \toprule
    \textbf{Model} & \multicolumn{4}{c}{Saliency} & Faithfulness\\
      \cmidrule(lr){2-5} \cmidrule(lr){6-6}
     & BLEU &METEOR & ROUGE-L &BERTScore & CLIP Score\\
    \midrule \\[-1.2em]
    \multicolumn{6}{l}{\textbf{Baseline Model}} \\
    MiniGPT4      &9.49    & 23.24   &15.83 &85.63 &  22.62    \\
    MiniGPT4-Woodpecker      &10.42    & 23.79   &16.66 &86.20 &  \textbf{22.99}    \\
    MiniGPT4-LURE      &10.19    & 20.39   &15.42 &85.47 &  22.71  \\\midrule \\[-1.2em]
    \multicolumn{6}{l}{\textbf{Our Model}} \\
   
    MiniGPT4-Pruned   &10.69   & 21.06    & 16.48   & 86.06  &  22.81   \\
    MiniGPT4-Appended &10.19    & \textbf{24.59}    & 16.23    & 86.15  &  22.74    \\
     MiniGPT4-Combined &\textbf{11.17}    & 22.51    & \textbf{16.84}    & \textbf{86.34}  &  22.96     \\ \midrule \\[-1.2em]
    \multicolumn{6}{l}{\textbf{Baseline Model}} \\
    LLaVA-1.5      & 13.89    & 24.79   &18.52 &87.47 &  23.14     \\
    LLaVA-1.5-Woodpecker      & 12.47    & 24.74   &18.20 &86.99 &  23.19     \\\hline \\[-0.99em]
    \multicolumn{6}{l}{\textbf{Our Model}} \\
    LLaVA-1.5-Pruned      & 13.90   & 21.55    & 18.48 &87.58 &  \textbf{23.34}     \\
    LLaVA-1.5-Appended      & 13.71   & \textbf{25.58}   & 18.43 &87.47 &  23.18    \\
     LLaVA-1.5-Combined      & \textbf{15.07}   & 22.84   & \textbf{18.58} &\textbf{87.61} &  \textbf{23.34 }   \\\bottomrule
  \end{tabular}
\end{table*}


\subsection{Ablation Studies}

To investigate the effect of our two trained critic models, we experiment on both datasets with two variants of our full method (i.e., -\textbf{Combined}): ``\textbf{-Pruned}'', which only removes hallucinated and non-salient features identified by our critic model; and ``\textbf{-Appended}'', which only appends missing salient image features generated by our critic model. 
As we see in Table~\ref{tab:HouseResult} and Table~\ref{tab:productResult}, both variants positively contribute to improving saliency and faithfulness.

We also assess our trained critic models' (based on fine-tuning BLIP-2 on our training data) performance with the non-fine-tuned BLIP-2 model at the feature-level. Table~\ref{tab:part-1} shows the feature classification accuracy of our trained critic model-$3a$ and non-fine-tuned BLIP-2 model on three types of image features: hallucinated, salient, and non-salient in the test set of the House data.
It is observed that although the non-fine-tuned BLIP-2 model achieves equal accuracy in identifying hallucinated features, its performance is significantly worse in identifying salient and non-salient features compared to our trained critic model.

\begin{table}[hbtp]
\small
  \centering
  \caption{Evaluation of image feature classification accuracy into hallucinated (Hal), salient (Sal) and non-salient (Non-Sal) labels on the House Dataset.}
  \label{tab:part-1}
  \begin{tabular}{lccc}
    \toprule \\ [-1.2em]
    \textbf{Model}& \textbf{Hal}  &\textbf{Sal} &\textbf{Non-Sal}\\
    \midrule \\[-1.2em]
    Trained BLIP2 Model-$3a$      & \textbf{96.12} &\textbf{92.93}    &\textbf{71.20}        \\
   Non-fine-tuned BLIP2 Model     & \textbf{96.12} & 57.32 &41.77         \\\bottomrule
  \end{tabular}
\end{table}

Our critic model-$3b$ generates a list of salient features from the input image. We measure the quality of the generated list of salient features in terms of saliency and faithfulness. Table~\ref{tab:part-2} shows the comparison between the two models. We measure the saliency of the generated features list by comparing this generated features list with the list of ground-truth salient features using Sentence-BERT (SBERT) similarity score  ~\cite{reimers2019sentenceSBERT}. For faithfulness, considering the images, we use  CLIPScore~\cite{hessel2021clipscore}. The SBERT score shown in Table~\ref{tab:part-2} shows that our model-generated salient features are more similar to ground-truth salient features compared to the salient features generated by the non-fine-tuned BLIP-2 model. The CLIPscore shown in Table~\ref{tab:part-2} shows the generated features are comparably-aligned with the input images.
\begin{table}[hbtp]
\small
  \centering
  \caption{Evaluation of generated salient features on the House Dataset.}
  \label{tab:part-2}
  \begin{tabular}{p{10em}cc}
    \toprule\\ [-1.2em]
    \textbf{Model}& \textbf{SBERT Score} &\textbf{CLIP Score} \\
    
    \midrule \\[-1.2em]
    Trained BLIP2 Model-$3b$      & \textbf{54.87}    & 27.05          \\
    Non-fine-tuned BLIP2 Model     & 45.01   &  \textbf{27.46 }       \\\bottomrule
  \end{tabular}
\end{table}

\section{Conclusion}
In this paper, we propose a novel approach to generating text that is both faithful and salient from mixed-modal data that includes images and structured data. To ensure salient and faithful text generation, we train a small vision critic model to: (i) identify the hallucinated, salient and non-salient features, and (ii) generate a list of salient features from images. This information is used in the \emph{post editing} step to improve generation quality. Experimental results on two mixed-modal datasets demonstrate that our framework outperforms recent large multimodal models as well techniques specifically designed to reduce hallucination in terms of faithfulness and saliency metrics.


\noindent\textbf{Limitation and Future work} 
Our critic model sometimes prunes  subjective features such as "Eye-catching", "Amazing opportunity", "Elegant beauty", "Piece of luxury" etc, which are essential for making the advertising text attractive. In future, we will consider this issue. In addition, we also plan to explore the saliency and hallucination problem in other modalities such as videos and audios.

\section*{Ethical Considerations}
Our model utilizes existing pre-trained vision language model, thus the ethical concerns associated with these models would also be applicable to our proposed framework.

\section*{Acknowledgments}
This material is based on research sponsored by Defense Advanced Research Projects Agency (DARPA) under agreement number HR0011-22-2-0047. The U.S. Government is authorised to reproduce and distribute reprints for Governmental purposes notwithstanding any copyright notation thereon. The views and conclusions contained herein are those of the authors and should not be interpreted as necessarily representing the official policies or endorsements, either expressed or implied, of DARPA or the U.S. Government.

\bibliography{acl_latex}
\clearpage
\appendix

\section{Prompt Template for Text Generation from LMM}
\label{app-prompt-design-gen}
\begin{figure*}[hbtp]
\vspace{-0.2cm}
\begin{center}
\includegraphics[width=0.95\textwidth]
{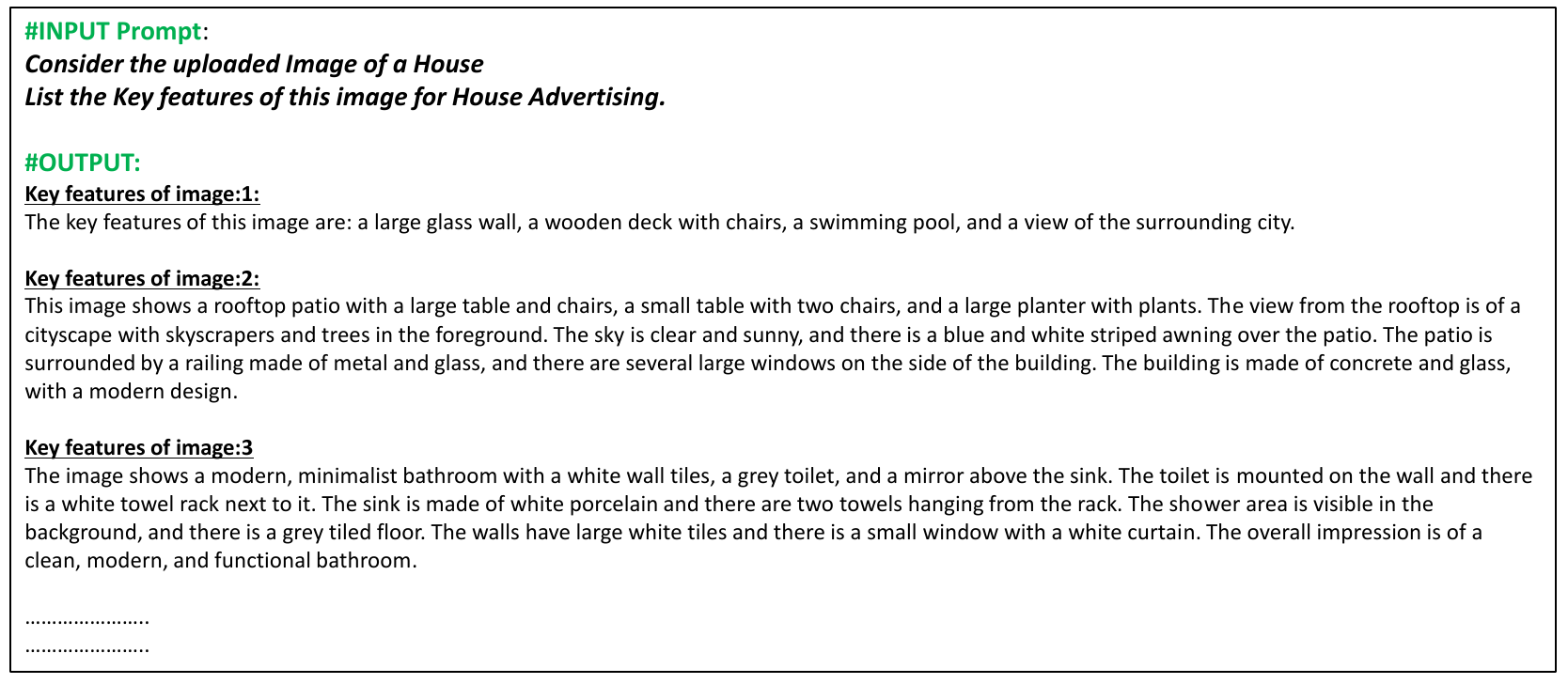}
\end{center}
\vspace{-0.2cm}
\caption{Prompt Template for LMM to generate key features of the image for House dataset} 
\label{fig:PromptImageKeyFeatures}
\end{figure*}

\begin{figure*}[hbtp]
\vspace{-0.2cm}
\begin{center}
\includegraphics[width=0.95\textwidth]
{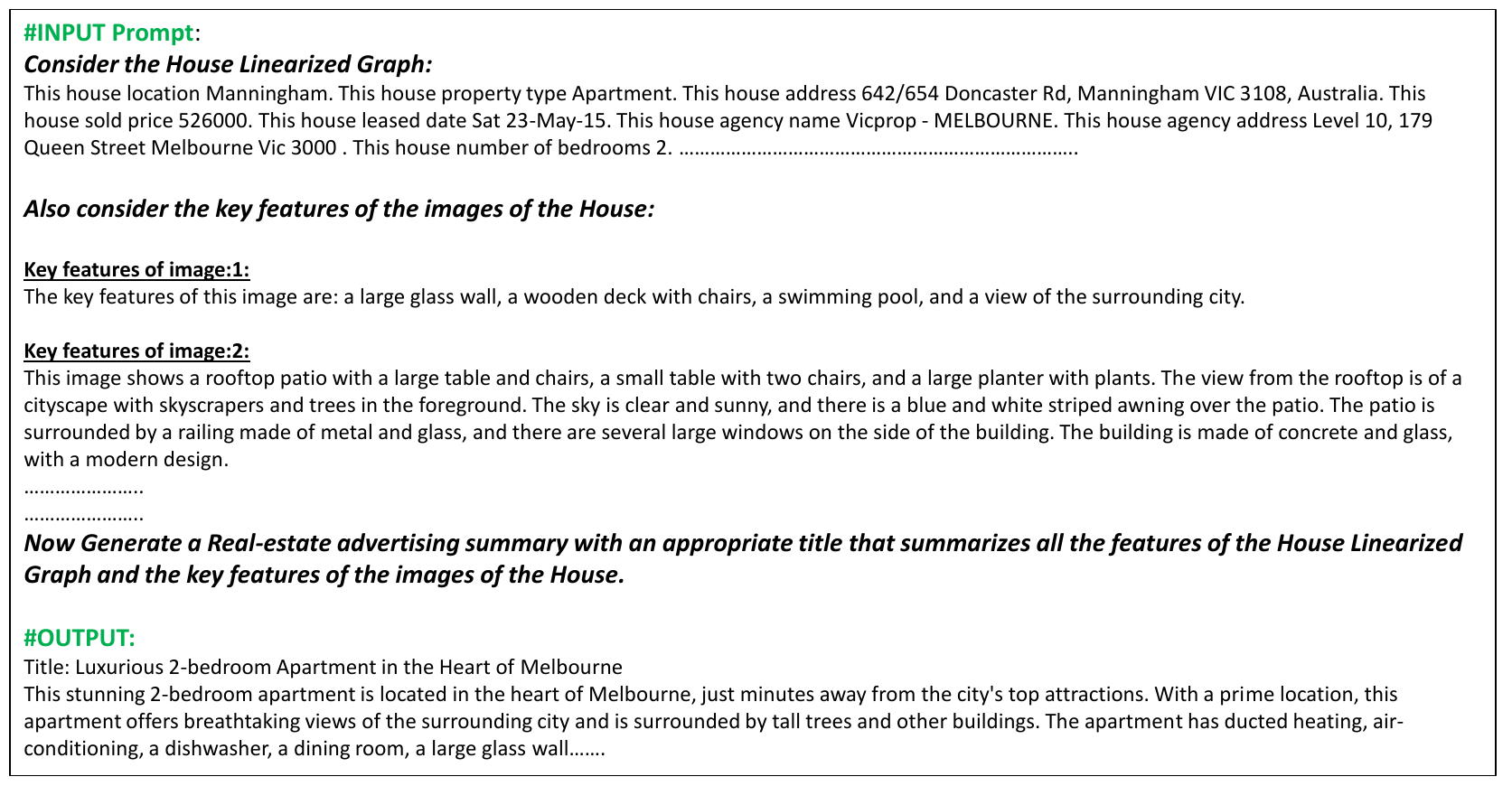}
\end{center}
\vspace{-0.2cm}
\caption{Prompt Template for LMM to generate text for House dataset} 
\label{fig:PromptTextGeneration}
\end{figure*}

The two prompt templates for House dataset are shown in Figure~\ref {fig:PromptImageKeyFeatures} and in Figure~\ref {fig:PromptTextGeneration} 

\section{Prompt Template for Training Data Generation}
\label{app-prompt-design-train}

\begin{figure*}[hbtp]
\vspace{-0.2cm}
\begin{center}
\includegraphics[width=0.85\textwidth]
{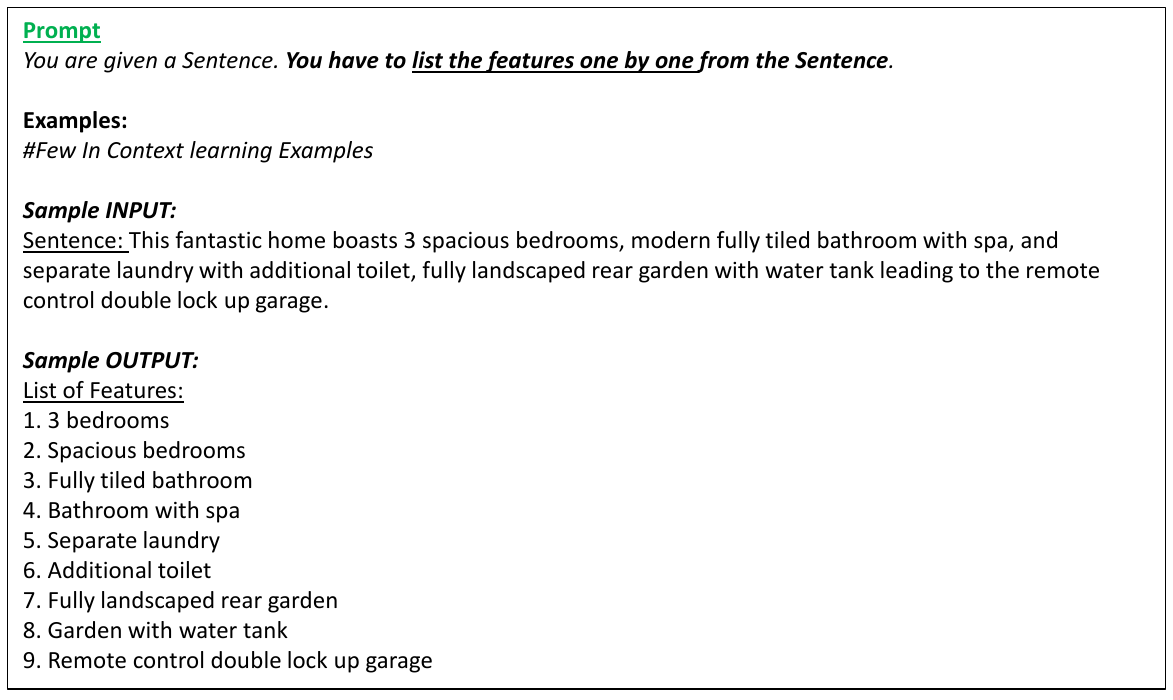}
\end{center}
\vspace{-0.2cm}
\caption{Prompt Template for LLM to extract list of features from a sentence} 
\label{fig:PromptListOfFeatures}
\end{figure*}

\begin{figure*}[hbtp]
\vspace{-0.2cm}
\begin{center}
\includegraphics[width=0.95\textwidth]
{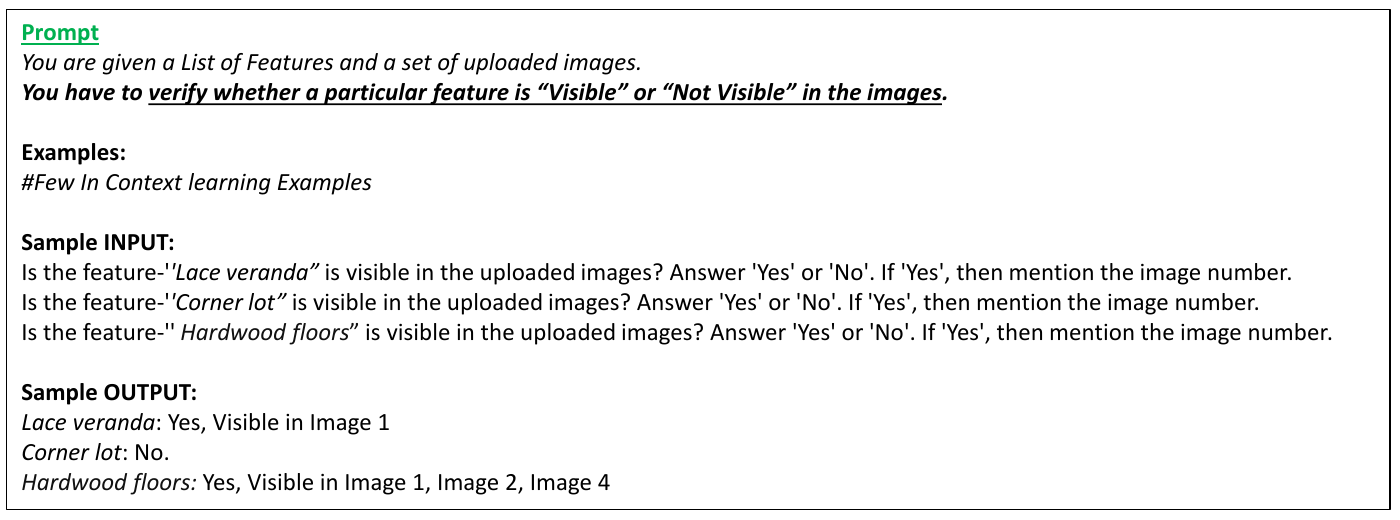}
\end{center}
\vspace{-0.2cm}
\caption{Prompt Template for GPT-4V to list visible and not visible image features} 
\label{fig:PromptvisibleFeatures}
\end{figure*}

\begin{figure*}[hbtp]
\vspace{-0.2cm}
\begin{center}
\includegraphics[width=0.95\textwidth]
{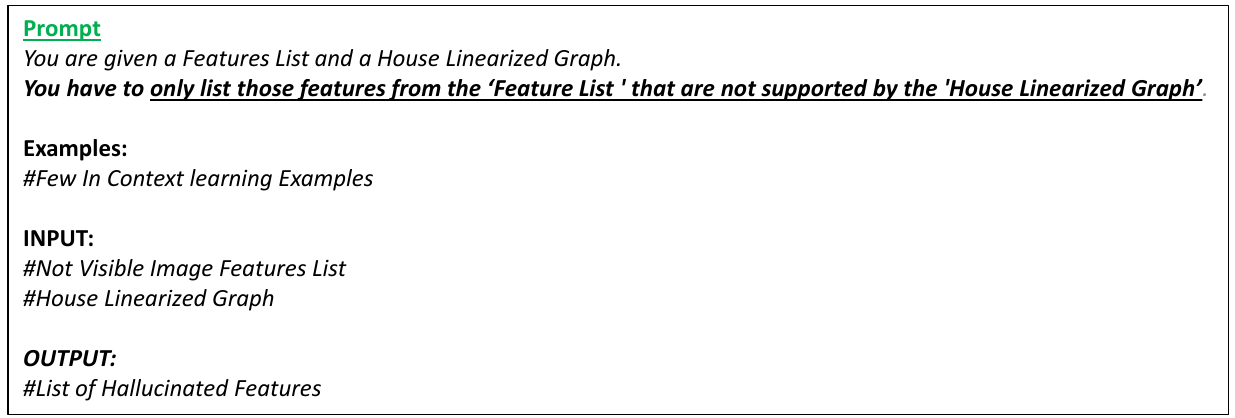}
\end{center}
\vspace{-0.2cm}
\caption{Prompt Template for LLM to list hallucinated features} 
\label{fig:PromptHalFeatures}
\end{figure*}

\begin{figure*}[hbtp]
\vspace{-0.2cm}
\begin{center}
\includegraphics[width=0.95\textwidth]
{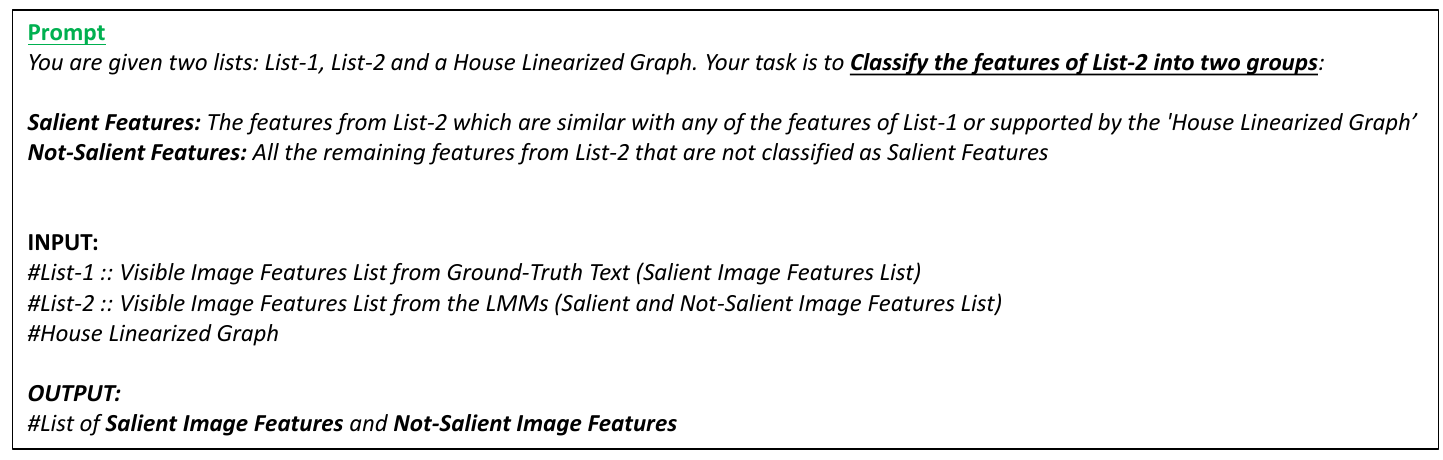}
\end{center}
\vspace{-0.2cm}
\caption{Prompt Template for LLM to list salient and not salient image features} 
\label{fig:PromptsalientFeatures}
\end{figure*}

\begin{figure*}[hbtp]
\vspace{-0.2cm}
\begin{center}
\includegraphics[width=0.95\textwidth]
{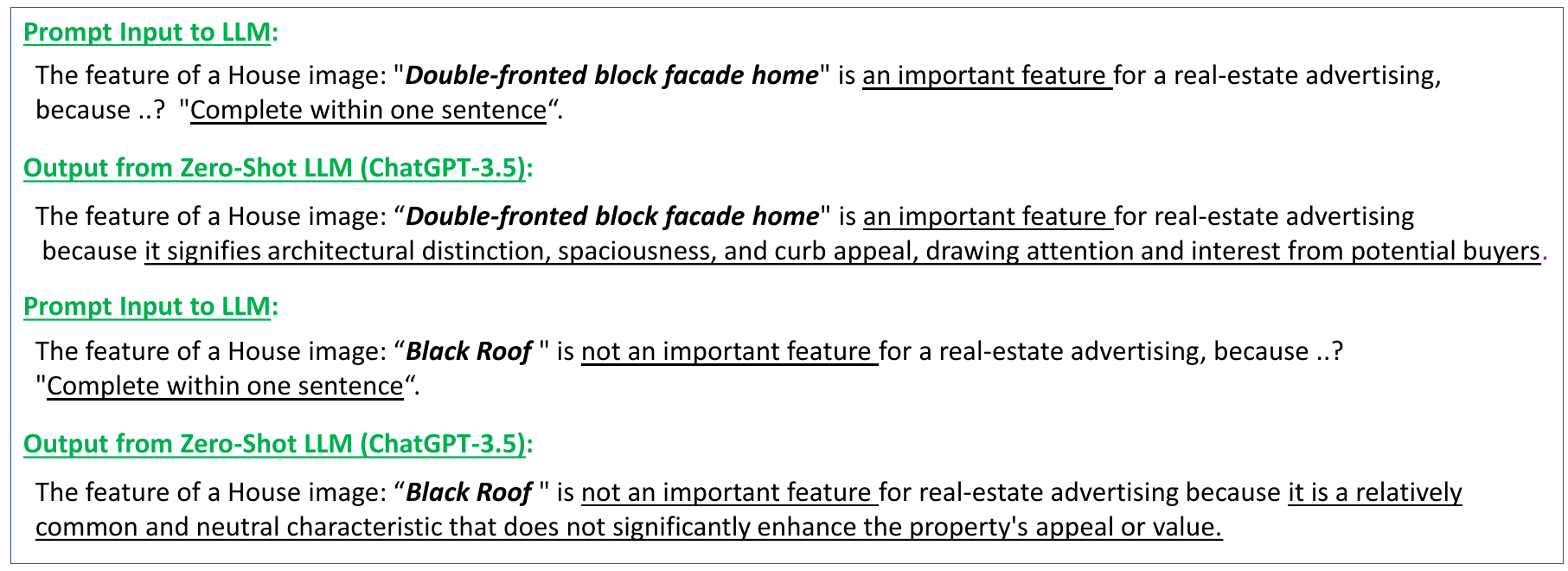}
\end{center}
\vspace{-0.2cm}
\caption{Prompt Template for LLM to generate rationale for salient features and not-salient features} 
\label{fig:PromptRationale}
\end{figure*}

\textbf{(1) Extracting features from text:}
Prompt template for Extracting features from the sentence of the text is shown in Figure:~\ref{fig:PromptListOfFeatures}. \\
\textbf{(2) Listing visible and non-visible features:}
Prompt template for listing visible and not visible features from the list of features is shown in Figure:~\ref{fig:PromptvisibleFeatures}.\\
\textbf{(3) Listing hallucinated features:}
Prompt template for listing hallucinated features is shown in Figure:~\ref{fig:PromptHalFeatures}.\\
\textbf{(4) Listing salient and non-salient features:}
Prompt template for listing salient and not salient features is shown in  Figure:~\ref{fig:PromptsalientFeatures}.\\
\textbf{(5) Generating rationale for features: }
The prompt template for generating rationale is shown in Figure~\ref{fig:PromptRationale}.

\section{Prompt Design for Post-hoc Text Editing using LLM}
\begin{figure*}[hbtp]
\vspace{-0.2cm}
\begin{center}
\includegraphics[width=0.7\textwidth]
{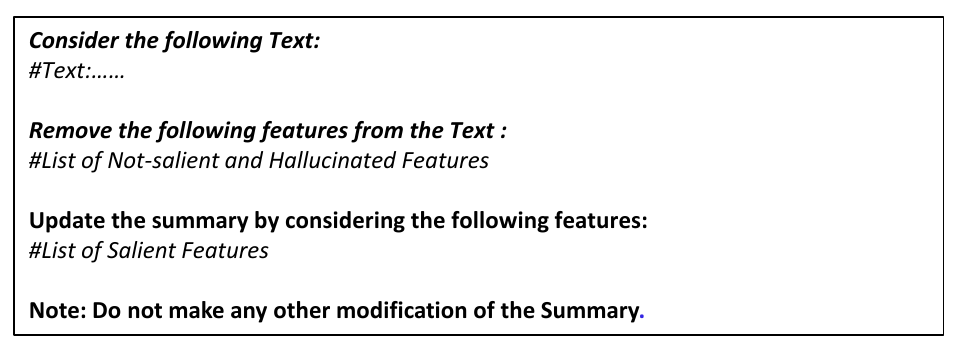}
\end{center}
\vspace{-0.2cm}
\caption{Prompt Template for LLM to do post-hoc correction of the generated text according to the feedback of the critic model} 
\label{fig:PromptUpdatesummary}
\end{figure*}

The prompt template is shown in Figure:~\ref{fig:PromptUpdatesummary} for updating the text using LLM GPT-3.5 according to the feedback of the critic model.

\section{Prompt Design for Preparing Salient and Faithful Ground-truth Text}
\begin{figure*}[hbtp]
\vspace{-0.2cm}
\begin{center}
\includegraphics[width=0.99\textwidth]
{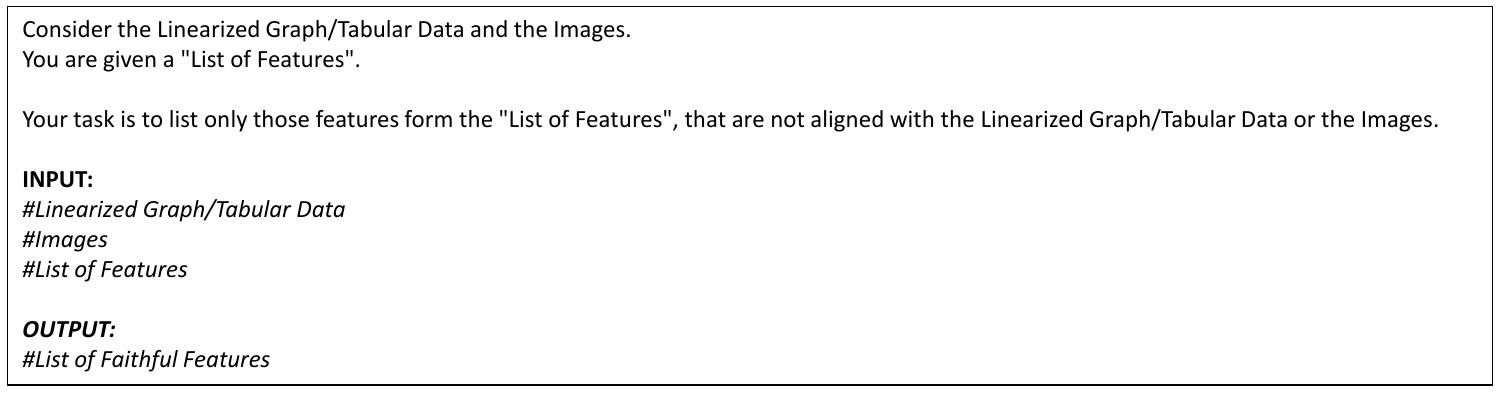}
\end{center}
\vspace{-0.2cm}
\caption{Prompt Template for GPT-4V to list faithful and salient features from the ground-truth text} 
\label{fig:PromptFaithFeaturesPara}
\end{figure*}

\begin{figure*}[hbtp]
\vspace{-0.2cm}
\begin{center}
\includegraphics[width=0.99\textwidth]
{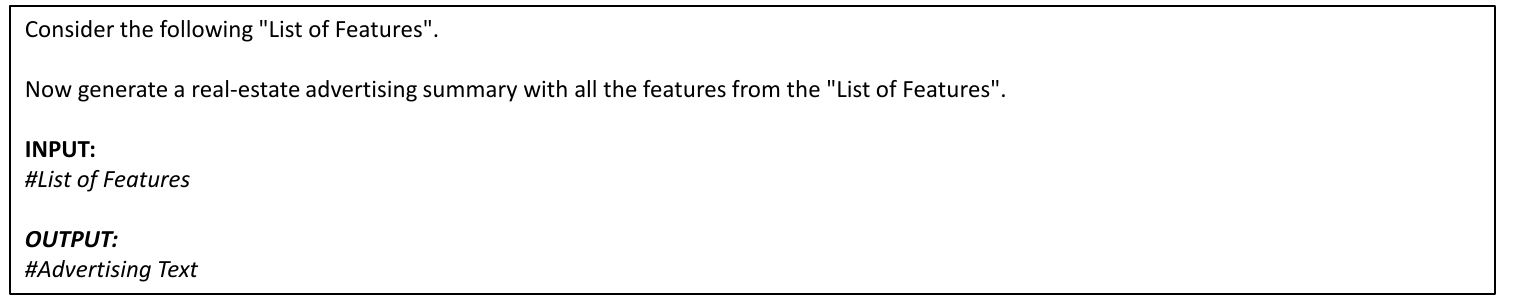}
\end{center}
\vspace{-0.2cm}
\caption{Prompt Template for LLM to make a paragraph with the faithful and salient features} 
\label{fig:PromptPara}
\end{figure*}

We use the prompt in Figure:~\ref{fig:PromptFaithFeaturesPara} to extract the faithful and salient features from the hallucinated ground-truth text. Then using the prompt in FIgure:~\ref{fig:PromptPara}, we generate a final salient and faithful ground-truth text. 

\section{Data Split Ratio for Training the Critic Model}\label{app:data-split}

\begin{table}[hbtp]
\small
  \centering
  \caption{House Data split ratio for Critic Model-$3a$}
  \label{tab:stat_house}
  \begin{tabular}{l|c}
  \hline
    \textbf{Data}& \textbf{\#Num. of Instances} \\\hline \\[-1em]
     \textbf{Training Samples} &\textbf{9517} \\
    --Hallucination Features     & 5478           \\
    --Salient Features     & 2703           \\
    --Not-Salient Features     & 1336          \\
    \midrule \\[-1.2em]
    \textbf{Validation Samples} &\textbf{1418} \\
    --Hallucination Features     & 845           \\
    --Salient Features     & 409           \\
    --Not-Salient Features     & 164          \\\bottomrule
  \end{tabular}
\end{table}

\begin{table}[hbtp]
\small
  \centering
  \caption{House Data split ratio for Critic Model-$3b$}
  \label{tab:stat-house-critic-3b}
  \begin{tabular}{l|c}
  \hline
    \textbf{Data}& \textbf{\#Num. of Instances} \\\hline \\[-1em]
     \textbf{Training Samples} &\textbf{10,359} \\
     \midrule \\[-1.2em]
    \textbf{Validation Samples} &\textbf{1155} \\
    \bottomrule
  \end{tabular}
\end{table}

\begin{table}[hbtp]
\small
  \centering
  \caption{Product Data split ratio for the Critic Model-$3a$}
  \label{tab:stat-product-critic-3a}
  \begin{tabular}{l|c}
  \hline
    \textbf{Data}& \textbf{\#Num. of Instances} \\\hline \\[-1em]
     \textbf{Training Samples} &\textbf{13202} \\
    --Hallucination Features     & 2529           \\
    --Salient Features     & 6795           \\
    --Not-Salient Features     & 3878       \\
    \midrule \\[-1.2em]
    \textbf{Validation Samples} &\textbf{2219} \\
    --Hallucination Features     & 393           \\
    --Salient Features     & 1243           \\
    --Not-Salient Features     & 583         \\\bottomrule
  \end{tabular}
\end{table}

\begin{table}[htpb]
\small
  \centering
  \caption{Product Data split ratio for Critic Model-$3b$}
  \label{tab:stat-product-critic-3b}
  \begin{tabular}{l|c}
  \hline
    \textbf{Data}& \textbf{\#Num. of Instances} \\\hline \\[-1em]
     \textbf{Training Samples} &\textbf{4235} \\
     \midrule \\[-1.2em]
    \textbf{Validation Samples} &\textbf{471} \\
    \bottomrule
  \end{tabular}
\end{table}

\section{Generated Samples}\label{sec:samples}
\begin{figure*}[htbp]
\tiny
\centering 
\begin{tabular}{|p{5.9in}|}
\hline 
\\  [-0.6em]
\small{\textbf{House Knowledge Graph and Images:}}  \\  [0.2em]
\hline 
  \begin{center}
      \includegraphics[scale=0.47]{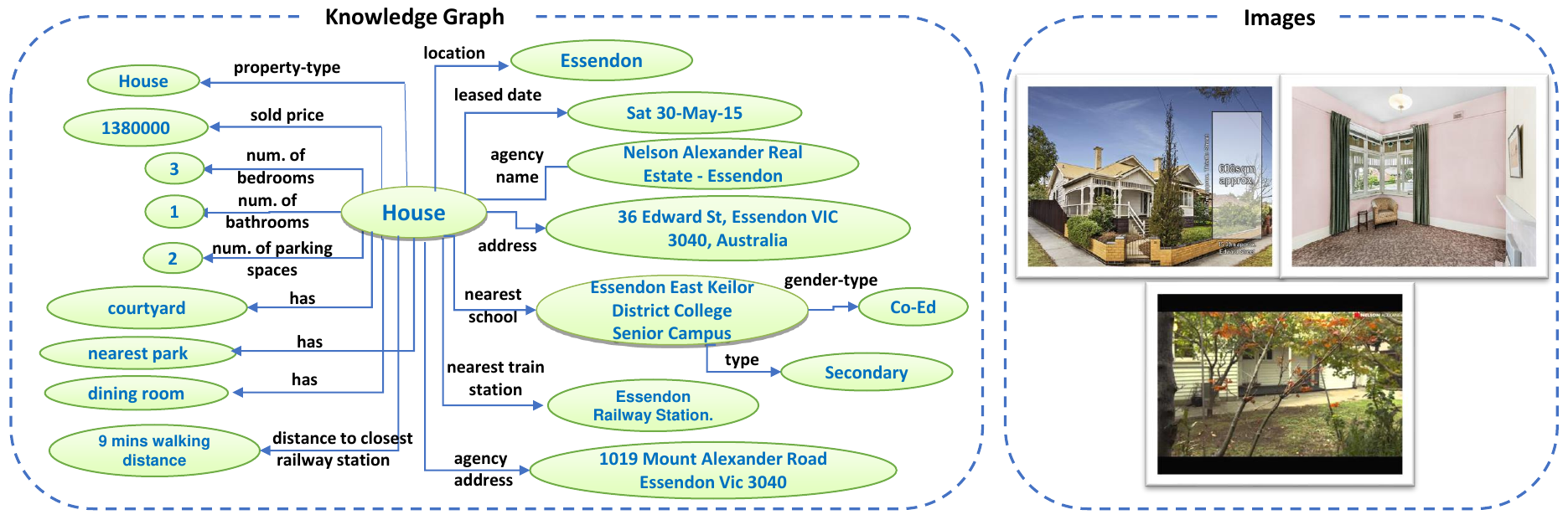}
  \end{center}\\[-0.9em]
\hline
\\[-0.5em]

\textbf{\underline{Pre-processed Ground-truth text:}} \\\\[-0.8em] 
\textbf{Stunning Edwardian Charm in Prestige Essendon Location!:} This grand home features three spacious bedrooms, a central bathroom, elegant formal lounge, dining room, and adjoining kitchen. Enjoy high ceilings, lead-light windows, a cellar, hallway storerooms, and a wrap-around verandah overlooking the expansive north-facing backyard. With a garage, carport, and walking distance to local parks and Essendon Station, this elevated corner block boasts a total land area of 603sqm. Don't miss out on this rare opportunity to own a piece of Edwardian history in one of Essendon's most sought-after neighborhoods!\\ [0.3em]

\hline
\\[-0.5em]
\textbf{\underline{Our model:}} \\\\[-0.8em] \textbf{Spacious 3-bedroom House in Essendon with Picket Fence and White Porch:} This advertisement is for a charming classical Victorian home in the urban area of Essendon, Melbourne. The house has a picket fence surrounding the front yard, a small porch with a white railing and two small windows, a small tree, and a small garden. The key features of the image are the yellow exterior and white trim, small garden, and white wooden porch with large windows and wooden frames. Located on a generous block, this charming home boasts original period features that add to its character. The spacious living room is perfect for families looking for a cozy, well-maintained, and well-located house in Essendon. Additional information provided includes the address (36 Edward St, Essendon VIC 3040, Australia), sold price (\$1380000), leased date (Sat 30-May-15), agency name (Nelson Alexander Real Estate - Essendon), agency address (1019 Mount Alexander Road Essendon Vic 3040), number of bedrooms (3), number of bathrooms (1), number of parking spaces (2), whether it has a courtyard, a dining room, a nearby park, the number of carport spaces (1), the number of garage spaces (1), proximity to the nearest train station (9 mins walking distance), and the nearest school (Essendon East Keilor District College Senior Campus). The image key features are also highlighted, making this property an ideal choice for those seeking a Victorian home with period charm in Essendon.\\  [0.3em]
\hline
\\[-0.5em]
\textbf{\underline{Woodpecker model:}} \\\\[-0.8em]
\textbf{Charming 3-Bedroom House in Essendon with Picket Fence and White Porch:} This advertisement is for a spacious 3-bedroom house located in the desirable area of Essendon, Melbourne. The house features a charming yellow exterior with white trim, giving it a warm and inviting feel. The front yard is enclosed by a classic picket fence, adding to the overall charm of the property.Upon entering the house, you are greeted by a small porch with a white railing, providing the perfect spot to relax and enjoy the surroundings. The porch is adorned with two large windows, allowing for plenty of natural light to fill the interior space.The house is well-maintained and in good condition, with a small garden and a tree adding to the overall appeal. The key features of the image include the yellow exterior with white trim, the small wooden front door, and the white wooden porch with large windows and wooden frames.In addition to the visual features, the House Linearized Graph provides important information about the property. The house is located at 36 Edward St, Essendon VIC 3040, Australia. It was sold for \$1380000 and was leased on Sat 30-May-15. The agency responsible for the property is Nelson Alexander Real Estate - Essendon, located at 1019 Mount Alexander Road Essendon Vic 3040. The house offers 3 bedrooms, 1 bathroom, and 2 parking spaces, including 1 carport and 1 garage. It also features a courtyard and a dining room. The nearest park is within walking distance, providing a convenient outdoor space for recreation.For families looking for a cozy and well-located house, this advertisement is perfect. The property offers a spacious interior, a picket fence, and a small garden, creating a welcoming and comfortable atmosphere. The proximity to Essendon Railway Station, which is just a 9-minute walk away, adds to the convenience of the location. The nearest school, Essendon East Keilor District College Senior Campus, is also within close proximity, making it an ideal choice for families with school-aged children.Don't miss out on the opportunity to make this charming house in Essendon your new home. Contact Nelson Alexander Real Estate - Essendon today to arrange a viewing.\\  [0.3em]
\hline
\\[-0.5em]
\textbf{\underline{LURE model:}} \\\\[-0.8em]
The house is a 3-bedroom house with 1 bathroom and 2 parking spaces, located in a quiet neighborhood with easy access to parks and public transportation. It has a beautiful outdoor living area and a modern kitchen and living room with plenty of storage space. The bedrooms are spacious and bright, and the house is equipped with ducted heating and cooling. The house is located in Essendon, just minutes away from local shops, cafes, and restaurants. The area has several parks and reserves, perfect for recreation. The location is just a few minutes' walk away from Essendon Railway Station, which is a major transport hub. The house is in close proximity to several schools, including Essendon East Keilor District College Senior Campus, which is a CoEd school. Overall, this house is a great option for those looking for a charming and convenient place to live.
\\  [0.3em]
\hline
\\[-0.5em]
\textbf{\underline{MiniGPT4 model:}} \\\\[-0.8em]
\textbf{Spacious 3-bedroom House in Essendon with Picket Fence and White Porch:} This advertisement is for a charming, small yellow house in the urban area of Essendon, Melbourne. The house has a picket fence surrounding the front yard, a small porch with a white railing and two small windows, a small tree, and a small garden. The house is well maintained and in good condition. The key features of the image are the yellow exterior and white trim, small garden, small wooden front door, and white wooden porch with large windows and wooden frames.The House Linearized Graph provides additional information about the property, including its address (36 Edward St, Essendon VIC 3040, Australia), sold price (\$1380000), leased date (Sat 30-May-15), agency name (Nelson Alexander Real Estate - Essendon), agency address (1019 Mount Alexander Road Essendon Vic 3040), number of bedrooms (3), number of bathrooms (1), number of parking spaces (2), whether it has a courtyard, a dining room, a nearby park, the number of carport spaces (1), the number of garage spaces (1), proximity to the nearest train station (9 mins walking distance), and the nearest school (Essendon East Keilor District College Senior Campus). The image key features are also highlighted.This advertisement is perfect for families looking for a cozy, well-maintained, and well-located house in Essendon with a picket fence and a small garden. \\ [0.3em]
\hline
\end{tabular}
\caption{Example of generated text on the House dataset. } 
\label{fig:houseExample}
\end{figure*}

\begin{figure*}[htbp]
\small
\centering 
\begin{tabular}{|p{5.9in}|}
\hline 
\\  [-0.6em]
\small{\textbf{Product Tabular Data and Image:}}  \\  [0.2em]
\hline 
  \begin{center}
      \includegraphics[scale=0.47]{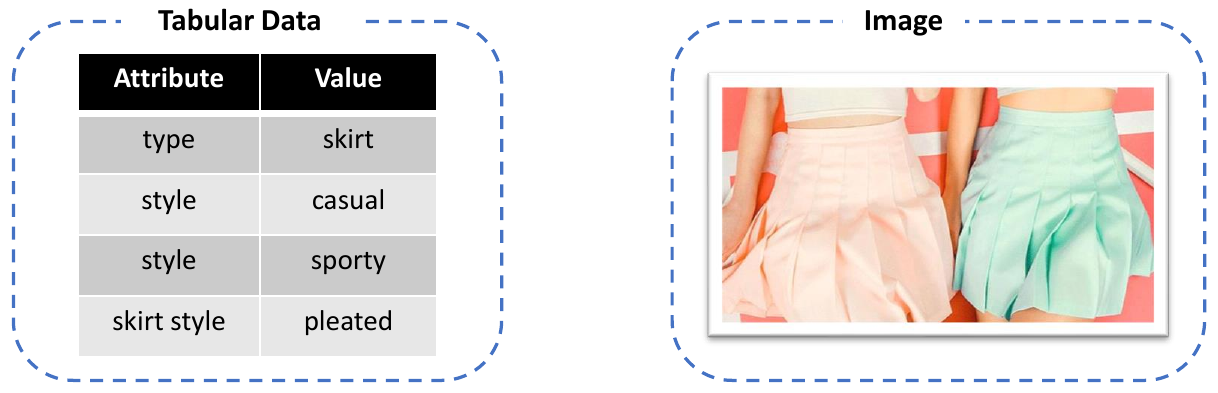}
  \end{center}\\[-0.9em]
\hline
\\[-0.5em]

\textbf{\underline{Pre-processed Ground-truth text:}} \\\\[-0.8em] 
Introducing our latest product that combines romantic and soft beauty with  a relaxed and neat style. This design highlights the integration of sporty and casual styles, making it popular with young girls. The dynamic layers of the pleated skirt hem add a hint of sweetness, while effectively embellishing the curve of the buttocks and highlighting the slim waist. Don't miss out on this stylish and flattering addition to your wardrobe!
\\ [0.3em]

\hline
\\[-0.5em]
\textbf{\underline{Our model:}} \\\\[-0.8em] The product is a pleated skirt with a slimming effect, featuring a structured triangular design at the hemline that reveals the lines of your legs. The four distinctive hem designs give the skirt a modern touch, making it a stylish and versatile addition to any wardrobe.
\\  [0.3em]
\hline
\\[-0.5em]
\textbf{\underline{Woodpecker model:}} \\\\[-0.8em]
Sample-1: The product is a loose-fitting silhouette outlining a hooded sweater style, showcasing a classic design that exudes a relaxed and energetic feeling. The color-blocking drawstring design with crystal patches adds decoration and captures attention. The fleece lining provides warmth, making you feel soft and cozy.
Sample-2: The product is a pleated skirt([pleated]) with a structured triangular design at the hemline that reveals the lines of your legs. The four distinctive hem designs give the skirt a modern touch. The product stands out with its unique shape, collar contrast color embroidery stitching, and striped element, which is timeless fashion.
Sample-3: The product is a collared shirt with a stand-up collar and unique shape. The shirt has a sporty style and is made of a
\\  [0.3em]
\hline
\\[-0.5em]
\textbf{\underline{LURE model:}} \\\\[-0.8em]
The image is of two women wearing pleated skirts. 
The first woman is wearing a mint green skirt, while the second woman is wearing a mint green skirt. Both skirts have a pleated design and a shorter hemline. The women are standing next to each other and smiling. The image is shot from the waist up, so only their heads and upper bodies are visible. The background has a repeating pattern of small circles. The setting is casual, outdoors, with a bright pink or red pigment used on the walls and floor.
\\  [0.3em]
\hline
\\[-0.5em]
\textbf{\underline{MiniGPT4 model:}} \\\\[-0.8em]
Sample-1: The product is a loose-fitting silhouette outlining a hooded sweater style, showcasing a classic design that exudes a relaxed and energetic feeling. The color-blocking drawstring design with crystal patches adds decoration and captures attention. The fleece lining provides warmth, making you feel soft and cozy.
Sample-2: The product is a pleated skirt with a structured triangular design at the hemline that reveals the lines of your legs. The four distinctive hem designs give the skirt a modern touch. The product stands out with its unique shape, collar contrast color embroidery stitching, and striped element, which is timeless fashion.
Sample-3: The product is a collared shirt with a stand-up collar and unique shape. The shirt has a sporty style and is made of a
 \\ [0.3em]
\hline
\end{tabular}
\caption{Example of generated text on the Product dataset. } 
\label{fig:product}
\end{figure*}

Figure~\ref{fig:houseExample} and Figure~\ref{fig:product} show qualitative examples of sample graph-images, sample tabular data-image, the pre-processed ground-truth texts, and the texts generated by different models on the House dataset and Product dataset, respectively.

\end{document}